\documentclass[conference]{IEEEtran}
\IEEEoverridecommandlockouts

\usepackage[T1]{fontenc}
\usepackage{cite}
\usepackage{amsmath,amssymb,amsfonts}
\usepackage{algorithmic}
\usepackage{graphicx}
\usepackage{textcomp}
\usepackage[table]{xcolor}
\definecolor{taskbg}{HTML}{EEF2F8}
\usepackage{booktabs}
\usepackage{float}
\usepackage{needspace}
\usepackage{multirow}

\begin{document}

\title{MaskHarness-WAM: Instance-Grounded Harnessing for Long-Horizon Robot Manipulation}

\author{
\IEEEauthorblockN{
Zitai Huang\textsuperscript{1,2,*},
Taiyi Su\textsuperscript{2,*},
Jian Zhu\textsuperscript{2,$\dagger$},
Jianjun Zhang\textsuperscript{1,2},
Chong Ma\textsuperscript{1,2}\\
Tianbin Liu\textsuperscript{2},
Weiyi Lu\textsuperscript{2},
Yi Xu\textsuperscript{2,$\ddagger$},
Hanli Wang\textsuperscript{1,$\ddagger$}
}
\IEEEauthorblockA{
\textsuperscript{1}Tongji University
\qquad
\textsuperscript{2}AIRC, Midea Group\\
\textsuperscript{*}Equal contribution \qquad
\textsuperscript{$\dagger$}Project Leader \qquad
\textsuperscript{$\ddagger$}Corresponding authors\\
Project Page: maskharness-wam.github.io
}
}

\maketitle

\begingroup
\renewcommand{\thefootnote}{}
\footnotetext{This work was completed while Zitai Huang was an intern at Midea AI
Research Centers.}
\addtocounter{footnote}{-1}
\endgroup

\begin{abstract}
Long-horizon robot manipulation requires not only stable local visuomotor control, but also continuous target tracking and reliable task progress assessment throughout execution. This challenge becomes particularly critical when multiple objects share identical appearances and must be manipulated in a prescribed order. In such scenarios, relying solely on a limited-horizon manipulation policy is often insufficient to determine which instance should be operated on and when the task should transition to the next stage. To address this challenge, we propose MaskHarness-WAM, an instance-grounded harness for long-horizon manipulation. The proposed system connects high-level task planning with low-level manipulation policies through target masks, while leveraging visual feedback for subtask scheduling and continuous execution. Since each subtask corresponds to a different target instance, the low-level policy requires a newly established initial target mask under the updated scene at each subtask transition. The harness continuously re-observes the environment, generates, and verifies the target mask at subtask boundaries, thereby updating the instance-level spatial condition provided to the low-level policy. Furthermore, the system advances the manipulation process by switching target instances according to the verified completion status of each subtask. Experiments on a real robot platform demonstrate that MaskHarness-WAM substantially outperforms limited-horizon policies on sequential multi-object manipulation, showing its effectiveness in extending local manipulation skills to reliable long-horizon execution.
\end{abstract}


\begin{figure*}[!t]
    \centering
    \includegraphics[width=\textwidth,page=1]{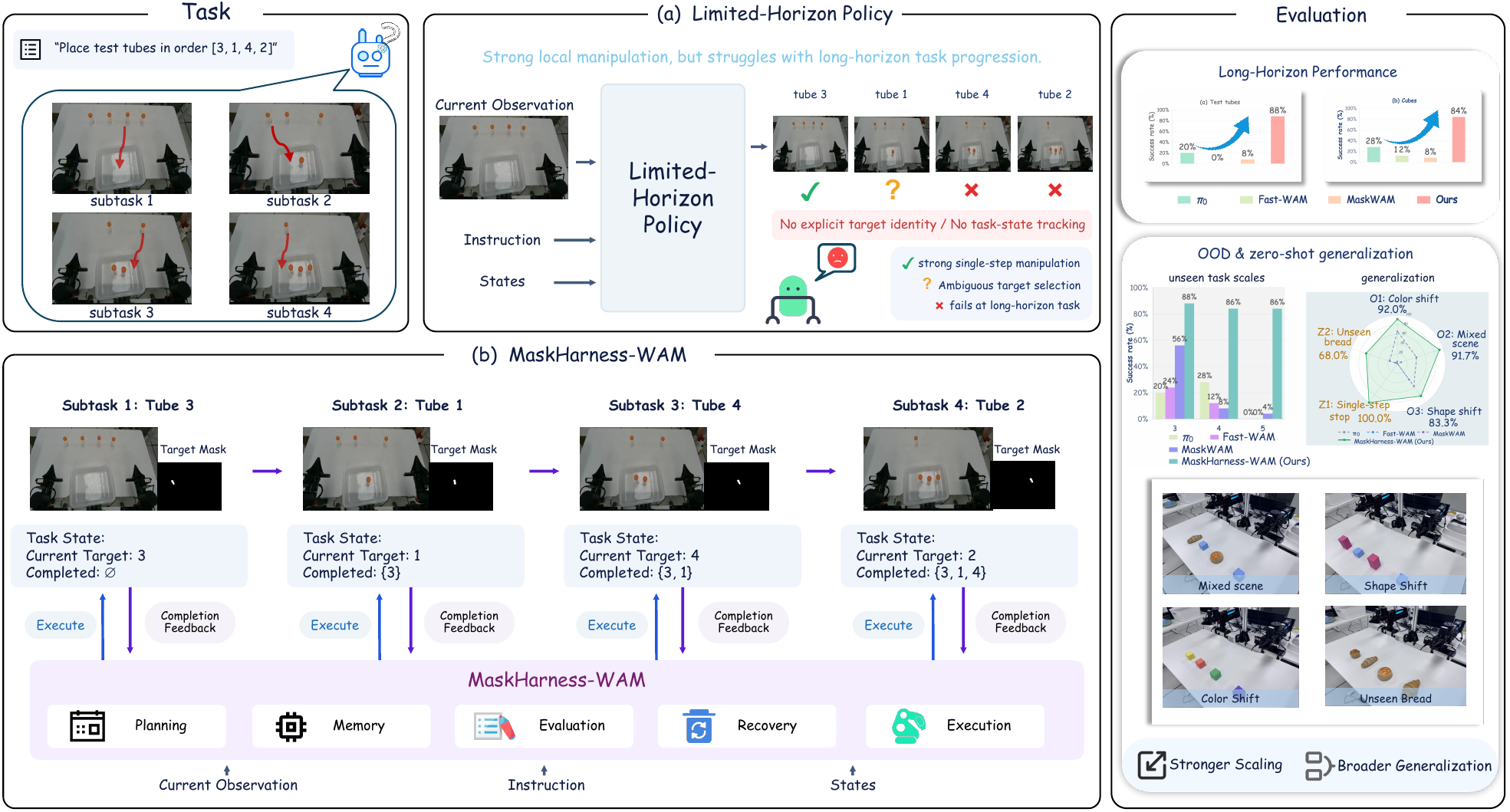}
    \caption{
    (a) Given a long-horizon instruction, limited-horizon policies (\emph{e.g.} $\pi_0$~\cite{black2025pi0} and Fast-WAM~\cite{yuan2026fastwam}) lack explicit instance identity and task-state tracking, leading to ambiguous target selection and failed stage transitions.
    (b) MaskHarness-WAM wraps the low-level policy with a task-level closed loop. It decomposes the instruction into ordered subtasks, maintains the current target and completed set, verifies a target mask before each execution, and advances only after completion is confirmed.
    (c) Evaluation shows improved long-horizon performance, together with stronger generalization under appearance shifts, scene-composition changes, geometry shifts, and zero-shot unseen-category transfer.}
    \label{fig:overview}
\end{figure*}

\section{Introduction}

Reliable execution of long-horizon instructions remains a fundamental goal toward general-purpose robot manipulation. In real-world scenarios such as industrial assembly, warehouse fulfillment, and laboratory automation, robots are often required to manipulate multiple objects with identical appearances in a prescribed order. For example, identical screws may need to be inserted into different holes sequentially, while visually indistinguishable test tubes must be transferred and inspected according to their sample identities. Although these objects share similar categories and visual appearances, they are not interchangeable within the task. Reliable execution of such procedures is therefore a critical step toward enabling robots to move beyond isolated skills and autonomously complete real-world workflows.

Recent advances in vision-language-action (VLA) models and world action models (WAM) have achieved significant progress in semantic understanding, action generation, and real-robot control~\cite{ye2026worldaction,yuan2026fastwam,motubrain2026,su2026demavla}. These models demonstrate strong capabilities on local manipulation skills, such as grasping and placement. However, most existing policies are designed for a limited execution horizon. Their reliability often depends on clear target specifications and a stable task stage. This assumption does not hold for sequential tasks involving multiple visually identical objects. Category-level semantics alone cannot determine the correct target instance. Moreover, completing a local motion does not necessarily indicate a valid transition to the next task stage. An incorrect target assignment can directly lead to improper actions, while an incorrect progress estimation may propagate through subsequent subtasks and cause cascading failures. Therefore, long-horizon sequential manipulation cannot be achieved by simply invoking local policies repeatedly. It requires an external management mechanism that maintains target identity, tracks task progress, and coordinates policy execution through continuous visual feedback.

Addressing the above challenges requires two complementary capabilities. First, an embodied harness introduces task planning, state maintenance, outcome evaluation, and failure recovery beyond the low-level policy, allowing finite-horizon manipulation skills to be organized into feedback-driven long-horizon behaviors~\cite{zhang2026harnessvla,gu2026harnesswam,huang2026roboharness,wang2026thea,ding2026zetta,wang2026shaper,chen2026showharness}. Second, visual prompts such as masks ground linguistic targets into specific pixel regions, providing explicit spatial anchors for multiple visually identical instances and reducing the ambiguity introduced by category-level language descriptions~\cite{stone2023moo,sundaresan2023kite,fang2024moka,huang2025roboground,li2025controlvla,yu2026maskwam}. However, long-horizon tasks cannot rely on a static visual prompt provided only at the beginning of execution. Once an instance is completed, both the scene state and the identity of the next target change. Therefore, the system must update the target mask for each subsequent subtask based on the current observation, ensuring that the visual condition remains consistent with the active manipulation instance. The harness not only determines how multiple local operations are organized and advanced, but also provides the runtime mechanism for updating visual conditions across subtasks.

We propose \emph{MaskHarness-WAM}, an instance-grounded harness for long-horizon sequential manipulation. The key idea is to augment a mask-conditioned world action model enhanced with instance-level grounding with an external closed-loop harness. MaskHarness-WAM consists of two tightly coupled components: a task-level harness and a mask-conditioned WAM policy with instance-level grounding. Given a language instruction, the harness decomposes the task into an ordered sequence of subtasks and maintains the execution state, including the active target instance and completed instances. Before each WAM invocation, the harness grounds the current target instance into an instance-level visual condition by generating and verifying a target mask from the current observation. Only verified masks are provided to the WAM, preventing incorrect target binding from propagating into action generation.

The key insight of MaskHarness-WAM is to treat the target mask as a dynamic instance-level interface between task management and manipulation execution. Unlike approaches that provide a single visual prompt for an entire task, our harness updates the target condition at every subtask transition. The completion of one subtask triggers a new target selection and mask generation for the next stage, allowing the same low-level policy to operate on different instances throughout a long-horizon sequence. Conditioned on the verified mask, multi-view observations, robot state, and local instruction, the low-level policy generates a finite-horizon action chunk and executes it on the robot. The resulting observations are fed back to the harness for completion evaluation and task-state updates. By continuously alternating between target grounding, policy execution, and state transition, MaskHarness-WAM converts finite-horizon manipulation skills into reliable long-horizon behaviors over multiple identical objects.

Our main contributions are summarized as follows:
\begin{itemize}
    \item We introduce \emph{MaskHarness-WAM}, a harness system that extends finite-horizon WAM policies to long-horizon sequential manipulation without modifying the underlying policy.
    \item We propose an instance-grounded coordination mechanism that dynamically updates and verifies subtask-level target masks, enabling reliable target switching and closed-loop task progression.
    \item We demonstrate the effectiveness of MaskHarness-WAM through extensive real-robot experiments, achieving task success rates of at least 84\% across the main long-horizon manipulation settings, together with strong out-of-distribution and zero-shot generalization.
\end{itemize}

\section{Related Work}

\subsection{World Action Models and Visual Prompting}

World action models incorporate visual dynamics priors into policy learning by jointly modeling future observations and robot actions, demonstrating strong capabilities in cross-task generalization, future-state representation, and action generation~\cite{du2023unipi,wu2023gr1,cheang2024gr2,bharadhwaj2024gen2act,zhou2024robodreamer,zhu2025uwm,zhang2025dreamvla,ye2026worldaction,yuan2026fastwam,motubrain2026,ma2026dit4dit,zhang2026learning}. Existing approaches mainly improve the internal capabilities of these models through advances in video prediction, action modeling, and inference efficiency. MaskWAM~\cite{yu2026maskwam} further introduces masks as both explicit inputs and prediction targets, where future-mask prediction provides semantic supervision over task-relevant regions and the first-frame mask establishes a precise spatial anchor for manipulation. Beyond world action models, structured visual prompts and intermediate representations have been widely explored to improve target grounding in robot manipulation. These representations include points, bounding boxes, and masks~\cite{stone2023moo,sundaresan2023kite,fang2024moka,huang2025roboground,li2025controlvla}, as well as keypoints~\cite{haldar2025pointpolicy,huang2024rekep}, trajectories~\cite{gu2023rttrajectory,bharadhwaj2024track2act}, object flows~\cite{wen2023atm}, and affordance representations~\cite{bahl2023vrb,nasiriany2025rtaffordance}. These studies demonstrate that structured visual conditions can improve target awareness and spatial grounding for local manipulation policies. Different from approaches that modify model architectures or training objectives, MaskHarness-WAM focuses on how an harness system can verify, maintain, and update visual conditions throughout long-horizon execution.

\subsection{Robot Task Planning and Harness Systems}

Early language-driven robot systems explored moving high-level reasoning beyond continuous control. Language models can generate executable programs that coordinate perception and control interfaces~\cite{liang2023codeaspolicies,singh2023progprompt,huang2023voxposer}, or select pretrained manipulation primitives and perform task planning with environmental feedback~\cite{ahn2022saycan,huang2022innermonologue,rana2023sayplan,zhu2026dswam}. Building upon these ideas, embodied harness systems emphasize the role of an external runtime layer that connects perception, reasoning, and execution through tool or skill interfaces, while maintaining long-horizon feedback loops through persistent state, outcome evaluation, and failure recovery. Recent works have investigated several aspects of harness-based robot execution, including primitive-level invocation of frozen VLAs or WAMs, orchestration and transition across heterogeneous policies, scene-aware context and action termination evaluation, runtime criticism and recovery, skill and context evolution, and semantic action interfaces~\cite{zhang2026harnessvla,gu2026harnesswam,huang2026roboharness,wang2026thea,ding2026zetta,wang2026shaper,chen2026showharness}. These studies collectively suggest that reliable long-horizon robot behavior depends not only on the capability of low-level policies, but also on how an external system organizes policy invocation and interprets execution outcomes. Following this harness paradigm, our work focuses on sequential manipulation of duplicate objects. MaskHarness-WAM addresses two key challenges in long-horizon manipulation: maintaining correct target identity and reliably advancing task progress across multiple execution stages.

\begin{figure*}[t]
    \centering
    \includegraphics[width=\textwidth,page=1]{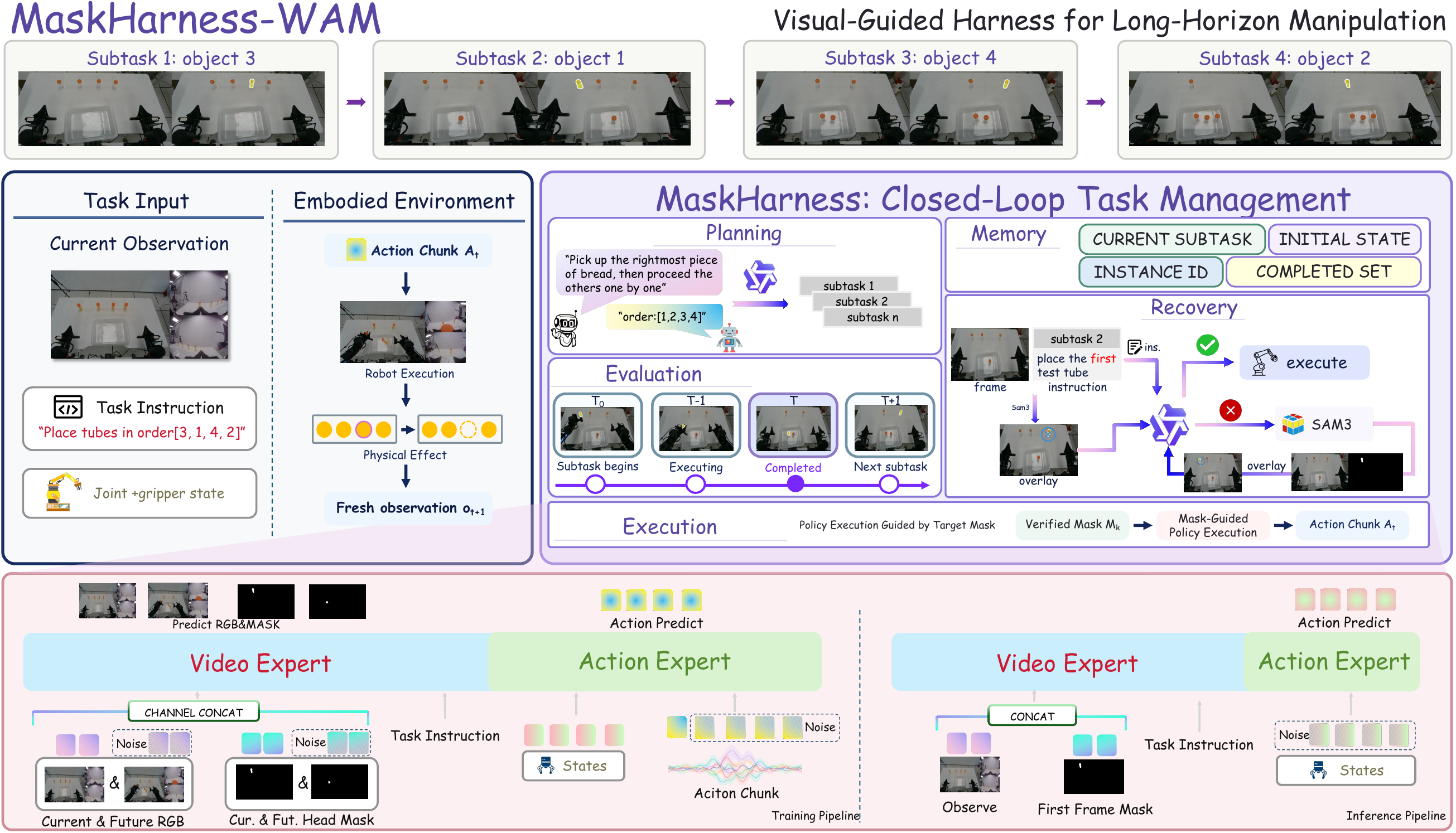}
    \caption{\textbf{Framework of MaskHarness-WAM.} Given a language instruction and current observations, the harness plans an ordered sequence of instance-specific subtasks and maintains the active target and execution history. At each subtask boundary, segmentation and visual verification establish a target mask for the low-level policy. Execution feedback determines whether to continue the current subtask or update the task state and switch targets; failed mask verification triggers re-observation and re-segmentation. The same policy is reused across successive targets (top). The lower panel shows joint RGB--mask--action training and action-only inference conditioned on the initial observation and target mask.}
    \label{fig:framework}
\end{figure*}

\section{Method}

\subsection{Task and Problem Formulation}

We study long-horizon robot manipulation tasks specified by complex language instructions, with a particular focus on sequential manipulation involving multiple duplicate objects. Here, duplicate objects refer to multiple instances belonging to the same object category that cannot be reliably distinguished solely by visual attributes such as color, size, or texture. Their identities must therefore be specified through spatial relations, such as ``the leftmost object'' or ``the third object from the left''. 

Unlike single-step manipulation, these tasks require the robot to continuously operate on multiple instances according to a prescribed order. At each stage, the system must ground the spatial description in the instruction to the correct object instance and determine whether the current operation has been completed based on scene changes. Failure to do so may result in missed targets, repeated operations, or premature stage transitions. Therefore, the problem jointly involves instance-level spatial grounding and cross-stage temporal management.

\subsection{Mask-Conditioned Policy}

MaskHarness-WAM uses a mask-conditioned world action model as the low-level execution primitive. The policy takes multi-view RGB observations, robot proprioceptive states, a local manipulation instruction, and a target mask, and outputs a finite-horizon bimanual action chunk $A_{1:H}\in\mathbb{R}^{H\times14}$, where $H$ denotes the action horizon and each action vector has 14 dimensions corresponding to the 7-DoF control of each of the two robot arms. Our focus is not to redesign the underlying WAM, but to enable its reliable reuse across long-horizon tasks by dynamically providing a verified instance-level target condition through the harness.

\smallskip \noindent\textbf{Mask Conditioning.}
The target mask serves as the visual grounding interface between the task-level harness and the low-level WAM. RGB observations and target masks are encoded with the same VAE, and their latent representations are concatenated along the channel dimension before being processed by the visual backbone. The mask is applied only to the head view, which provides a stable global view of all candidate objects, while the mask channels of the two wrist views are set to zero. This preserves the local geometric information from the wrist cameras while grounding target identity in a consistent global reference frame.

For each subtask, the harness supplies a verified target mask from the initial observation. The same low-level policy can therefore be re-grounded to different object instances as the task progresses. Importantly, the mask-conditioned policy itself remains unchanged across subtasks; instance switching is handled entirely by the harness through online mask generation, verification, and update.

\smallskip \noindent\textbf{Training.}
During training, the model jointly predicts robot actions, future RGB states, and future target masks. Future-mask prediction provides object-centric supervision that encourages the visual representation to preserve information about the active target throughout robot-object interaction. Target masks used for training are generated with SAM3~\cite{carion2026sam}. The overall training objective is
\begin{equation}
L=L_{\mathrm{rgb}}+L_{\mathrm{mask}}+L_{\mathrm{act}},
\end{equation}
where $L_{\mathrm{rgb}}$ supervises future visual prediction, $L_{\mathrm{mask}}$ supervises future target-mask prediction, and $L_{\mathrm{act}}$ optimizes action generation.

\smallskip \noindent\textbf{Inference.}
During real-robot execution, the policy directly predicts a finite-horizon action chunk from the current multi-view observations, proprioceptive state, local manipulation instruction, and the verified target mask from the initial frame. Future RGB observations and future masks are not explicitly generated at inference time. Instead, the target-aware representation learned through future-mask supervision allows a single initial mask to serve as an instance-level anchor throughout the action chunk. As a result, real-robot execution requires only one target mask per policy invocation, without per-frame segmentation or explicit target tracking.

\subsection{Harness System}

MaskHarness-WAM introduces a task-level closed loop around the low-level manipulation policy. The harness handles task decomposition, state tracking, visual evaluation, recovery, and execution scheduling, while continuous robot actions are generated entirely by the reusable low-level policy. Its role is to determine which subtask should be executed, which object instance should be grounded, and when the system should advance to the next stage. This separation allows a finite-horizon manipulation policy to be composed into feedback-driven long-horizon execution.

\smallskip \noindent\textbf{1) Planning.}
Given a language instruction, the planner decomposes the task into an ordered sequence of subtasks. Each subtask contains a local manipulation instruction and a target instance specified by a spatial relation, such as ``leftmost'' or ``third from the left.'' This decomposition separates the global execution order from individual manipulation operations, allowing the same low-level policy to be reused across different target instances.

\smallskip \noindent\textbf{2) Memory.}
The harness maintains a compact execution state across subtask boundaries, including the initial state, current subtask, target instance ID, and the set of completed instances. The initial state records the visual observation at the beginning of the active subtask, while the instance ID identifies the current manipulation target. The harness additionally retains the verified target mask associated with the active subtask. Together, these states connect the language-specified task order with online visual observations and provide consistent context for target verification, completion assessment, and task transitions.

\smallskip \noindent\textbf{3) Evaluation.}
Each subtask is controlled by two visual gates: an \emph{entry gate} for target verification and an \emph{exit gate} for completion assessment. Both gates are implemented with a vision-language model.

At the entry gate, the candidate target mask is overlaid on the full head-view image. The evaluator checks whether the highlighted region corresponds to the target instance specified by the current subtask. Only a verified mask is passed to the low-level policy, preventing incorrect instance grounding from propagating into action generation.

At the exit gate, the evaluator compares the current observation with the visual reference captured at the beginning of the subtask and determines whether the expected manipulation outcome has been achieved. If completion is not confirmed, the current subtask remains active. Once completion is verified, the harness marks the current instance as completed and activates the next subtask. The task terminates after all subtasks have been completed. The two gates therefore answer two complementary questions: \emph{which instance should be manipulated} and \emph{when should the task advance}.

\smallskip \noindent\textbf{4) Recovery.}
Target segmentation may fail because of occlusion, illumination changes, or densely arranged duplicate objects. If the entry gate rejects a candidate mask, the harness discards it and keeps the current task state unchanged. It then re-observes the scene, regenerates the target mask, and verifies it again before invoking the low-level policy. This recovery loop prevents incorrect target conditions from reaching the robot controller.

The same conservative strategy is applied to task progression. If the exit gate cannot verify completion, the harness remains at the current subtask and gathers new visual evidence rather than advancing prematurely. Recovery therefore protects both instance grounding and cross-stage task transitions.

\smallskip \noindent\textbf{5) Execution.}
Once the target mask is verified, the harness invokes the low-level policy with the current observation, robot state, local instruction, and target mask. The policy generates and executes a finite-horizon action chunk, after which the resulting observation is returned to the harness for completion evaluation. This forms a closed-loop process of target grounding, local execution, visual evaluation, and task transition, enabling the same low-level policy to support long-horizon sequential manipulation.

\section{Experiments}
\raggedbottom
\subsection{Experimental Setup}

\smallskip \noindent\textbf{Tasks.}
We evaluate MaskHarness-WAM on a real ARX dual-arm 7-DoF robotic platform. The evaluation focuses on two sequential manipulation tasks: \textbf{1) block} and \textbf{2) test-tube}. For both tasks, the training set contains only 4-instance demonstrations. Each 4-instance task admits 24 possible execution orders. The training set covers 18 orders for blocks and 15 orders for test tubes, while the remaining orders are reserved for evaluation.

At test time, we evaluate 3-, 4-, and 5-instance settings for both tasks, for a total of 6 sequential manipulation settings. Each setting is evaluated over 25 independent trials. The robot must follow the language-specified execution order and select the correct target among visually similar instances. The 3- and 5-instance settings evaluate generalization to task scales and execution horizons unseen during training. The 4-instance setting preserves the training task scale but uses unseen execution orders to evaluate generalization to novel task compositions. We further evaluate robustness under distribution shifts in object appearance, scene composition, and geometry. Finally, we evaluate zero-shot generalization to unseen tasks that are absent from the training set, without any task-specific demonstrations or additional fine-tuning.

\smallskip \noindent\textbf{Baselines.}
We compare MaskHarness-WAM with $\boldsymbol{\pi_0}$~\cite{black2025pi0}, \textbf{Fast-WAM}~\cite{yuan2026fastwam}, and \textbf{MaskWAM}~\cite{yu2026maskwam}, which represent general VLA policies, world action models, and mask-conditioned WAMs, respectively. For $\pi_0$ and Fast-WAM, we use their official open-source implementations. Since MaskWAM had not released its code at the time of our experiments, we reimplemented the method following the details provided in the original paper. All methods use the same robotic platform, RGB camera observations, initial scene distributions, and language instructions, and are evaluated under the same test settings.

\smallskip \noindent\textbf{Training and Inference Details.}
Each training segment contains 33 robot time steps, corresponding to an action prediction horizon of \(H=32\). The model is trained with a continuous flow matching objective, using 1,000 flow matching time steps and a time shift parameter of 5.0. We optimize the model with AdamW, using an initial learning rate of \(1\times10^{-4}\), weight decay of 0.01, cosine learning rate scheduling, bfloat16 mixed-precision training, and gradient clipping with a maximum norm of 1.0. Training is performed on 8 NVIDIA H20 96GB GPUs.

During inference, we use 10 denoising steps, running on a single NVIDIA RTX 5090 32GB GPU. Target masks are generated by SAM3~\cite{carion2026sam}. The mask verification and subtask completion evaluation modules in the harness are both powered by Qwen3.5~\cite{yang2025qwen3}. The two evaluation tasks use the same Qwen3.5, with different visual evidence and prompts.

\smallskip \noindent\textbf{Metrics.}
We use Success Rate (SR) and Progress Rate (PR) as evaluation metrics.
A trial is counted as successful only when all subtasks are completed in the specified order.
For the $i$-th trial, let $K_i$ denote the number of subtasks and $m_i$ denote the length of the longest correct subtask prefix starting from the beginning of the sequence. The two metrics are defined as
\begin{equation}
    \mathrm{SR} = \frac{1}{N}\sum_{i=1}^{N}\mathbb{I}\!\left[m_i = K_i\right],
    \qquad
    \mathrm{PR} = \frac{1}{N}\sum_{i=1}^{N}\frac{m_i}{K_i}.
\end{equation}

SR measures full-task success, while PR measures the fraction of subtasks completed in the correct order. Progress is defined by the longest correct prefix, so out-of-order operations and subtasks after the first error are not counted.

\subsection{Main Results}

Table~I reports quantitative comparisons among the four methods across different instance scales. 
The block tasks focus on target selection and sequential control among visually similar instances, while the test tube tasks further evaluate these capabilities under a different object configuration with elongated geometries and tighter placement tolerances. 
Figure~3 shows representative real-robot execution examples of both tasks.

\begin{figure*}[t]
    \centering
    \includegraphics[width=\textwidth,page=1,pagebox=cropbox]{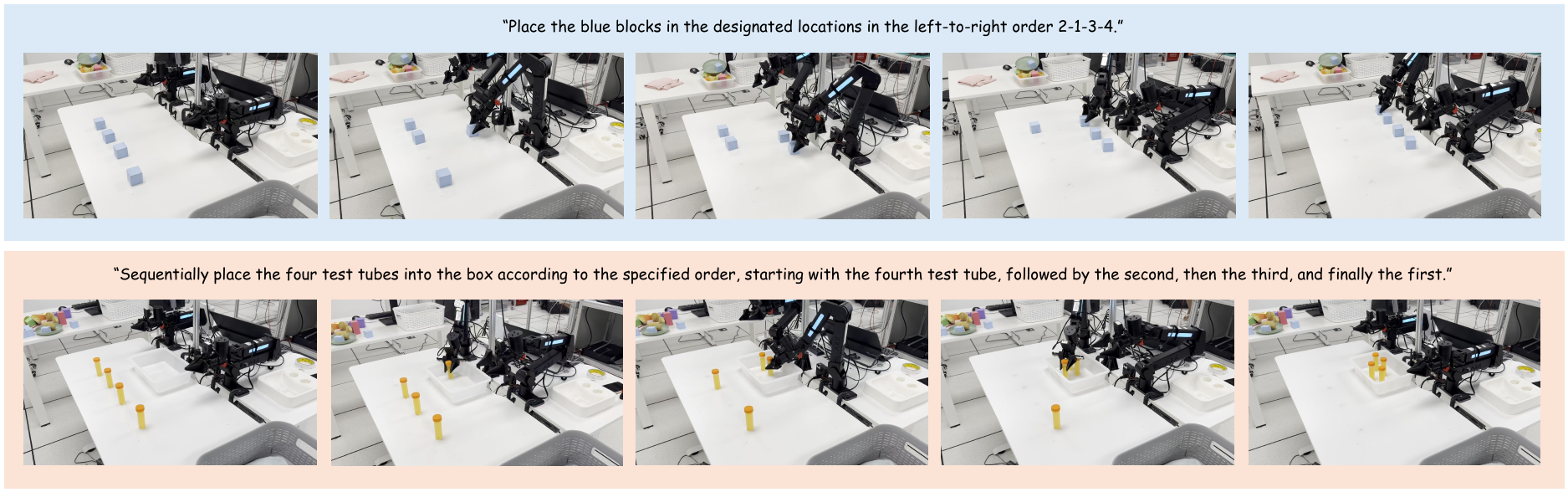}
    \caption{Real-robot sequential manipulation. Top: four blue blocks are placed in the specified left-to-right index order 2--1--3--4. Bottom: four test tubes are placed into the box in the order 4--2--3--1. Each row shows successive frames from one execution, illustrating target switching and task progression under MaskHarness-WAM.}
    \label{fig:main_rollouts}
\end{figure*}

\begin{table}[htbp]
    \centering
    \caption{Sequential manipulation results (\%).}
    \label{tab:main_results}
    \setlength{\tabcolsep}{3.0pt}
    \footnotesize
    \begin{tabular*}{\columnwidth}{@{\extracolsep{\fill}}lcccccc@{}}
        \toprule
        \multirow{2}{*}{\textbf{Method}} 
        & \multicolumn{2}{c}{\textbf{3 Instances}} 
        & \multicolumn{2}{c}{\textbf{4 Instances}} 
        & \multicolumn{2}{c}{\textbf{5 Instances}} \\
        \cmidrule(lr){2-3}\cmidrule(lr){4-5}\cmidrule(lr){6-7}
        & \textbf{SR} & \textbf{PR} 
        & \textbf{SR} & \textbf{PR} 
        & \textbf{SR} & \textbf{PR} \\
        \midrule

        \multicolumn{7}{>{\columncolor{taskbg}}c}{\textit{Cubes}} \\
        $\pi_0$~\cite{black2025pi0}     & 20.0 & 32.0 & 28.0 & 32.0 & 0.0 & 8.8 \\
        Fast-WAM~\cite{yuan2026fastwam} & 24.0 & 28.0 & 12.0 & 16.0 & 0.0 & 4.0 \\
        MaskWAM~\cite{yu2026maskwam}    & 56.0 & 72.0 & 8.0 & 31.0 & 4.0 & 23.2 \\
        \textbf{MaskHarness-WAM}         & \textbf{88.0} & \textbf{93.3} & \textbf{84.0} & \textbf{92.0} & \textbf{84.0} & \textbf{91.2} \\

        \addlinespace[2pt]

        \multicolumn{7}{>{\columncolor{taskbg}}c}{\textit{Test Tubes}} \\
        $\pi_0$~\cite{black2025pi0}     & 60.0 & 65.3 & 20.0 & 43.0 & 0.0 & 6.4 \\
        Fast-WAM~\cite{yuan2026fastwam} & 16.0 & 21.3 & 0.0 & 3.0 & 4.0 & 5.6 \\
        MaskWAM~\cite{yu2026maskwam}    & 8.0 & 45.3 & 8.0 & 42.0 & 0.0 & 21.6 \\
        \textbf{MaskHarness-WAM}         & \textbf{96.0} & \textbf{97.3} & \textbf{88.0} & \textbf{94.0} & \textbf{88.0} & \textbf{96.0} \\
        \bottomrule
    \end{tabular*}
\end{table}

Across all six settings, MaskHarness-WAM achieves the highest SR and PR while maintaining stable performance across different instance scales. For block tasks, SR reaches 88.0\%, 84.0\%, and 84.0\% under the 3-, 4-, and 5-instance settings, respectively; for test tube tasks, SR reaches 96.0\%, 88.0\%, and 88.0\%. The lowest PR across the six settings is 91.2\%. Performance therefore holds up under both task-scale variations and unseen execution orders.

Fast-WAM and MaskWAM degrade clearly across task settings. Even in the four-instance setting, which matches the training task scale, their full-task success rates fall far below MaskHarness-WAM. Both methods achieve relatively high PR in some settings, which indicates that their low-level manipulation remains effective. Their failures instead come from target selection and task progression. Without an explicit mechanism for task-state maintenance and target-condition updating, they struggle to follow the specified order among visually similar instances.

$\pi_0$ follows the specified order reasonably well in settings close to the training distribution. Its performance drops under unseen execution orders and varying instance scales, which indicates that its long-horizon generalization remains limited by the training distribution.

\subsection{Out-of-Distribution Generalization}

We further evaluate MaskHarness-WAM under out-of-distribution (OOD) visual conditions, where the task structure is preserved but the visual distribution differs from training. We consider three OOD settings. 
(1) Appearance Shift: the robot sequentially manipulates blocks with unseen colors, isolating changes in target appearance. 
(2) Scene Composition Shift: the robot manipulates bread objects in a mixed scene containing two blocks and two bread instances, introducing an unseen category composition together with distractors. 
(3) Appearance-and-Geometry Shift: the robot sequentially manipulates red cuboids among blue block distractors, jointly changing target color and geometry. 

\begin{figure*}[t]
    \centering
    \includegraphics[width=\textwidth,page=1,pagebox=cropbox]{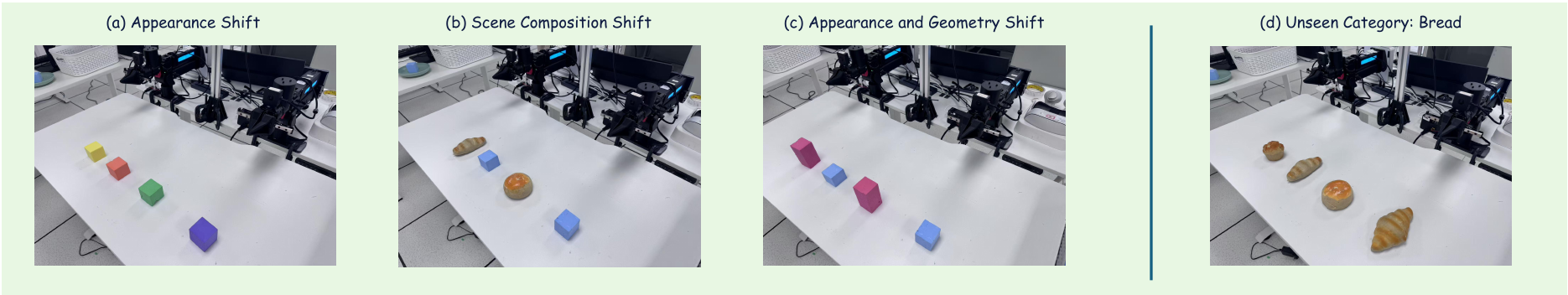}
    \caption{OOD and zero-shot generalization settings. 
    (a) Appearance Shift: sequential manipulation of blocks with unseen colors.
    (b) Scene Composition Shift: sequential manipulation of bread objects in a mixed scene with block distractors.
    (c) Appearance and Geometry Shift: sequential manipulation of red cuboids among blue block distractors. 
    (d) Zero-Shot Category Transfer: sequential manipulation of four bread instances without additional training on bread data.}
    \label{fig:generalization_scenes}
\end{figure*}

The Appearance Shift setting is evaluated with 25 trials per method, while the other two settings use 12 trials per method. SR and PR are computed according to Eq.~(2). In Table~\ref{tab:ood_results}, each entry is reported in the form SR/PR, where the value before the slash denotes SR and the value after the slash denotes PR. Representative scenes are shown in Fig.~\ref{fig:generalization_scenes}(a)--(c), and quantitative results are reported in Table~\ref{tab:ood_results}.

\begin{table}[htbp]
    \centering
    \caption{OOD generalization under three types of visual distribution shift (SR/PR, \%).}
    \label{tab:ood_results}
    \setlength{\tabcolsep}{2pt}
    \scriptsize
    \resizebox{\columnwidth}{!}{%
    \begin{tabular}{lcccc}
        \toprule
        \textbf{OOD Setting} & \textbf{$\boldsymbol{\pi_0}$} & \textbf{Fast-WAM} & \textbf{MaskWAM} & \textbf{MaskHarness-WAM} \\
        \midrule
        \textbf{Appearance Shift} 
        & 4.2/14.6 
        & 4.2/11.5 
        & 66.7/80.2 
        & \textbf{91.7/96.9} \\
        
        \textbf{Scene Composition Shift} 
        & 8.3/25.0 
        & 0.0/12.5 
        & 41.7/70.8 
        & \textbf{91.7/95.8} \\
        
        \textbf{Appearance-and-Geometry Shift} 
        & 0.0/16.7 
        & 41.7/54.2 
        & 58.3/79.2 
        & \textbf{83.3/83.3} \\
        \bottomrule
    \end{tabular}}
\end{table}

MaskHarness-WAM performs consistently well across all three OOD settings. Under Appearance Shift, it achieves 92.0\% SR and 97.0\% PR, showing that the instance-level mask condition remains effective when target appearance changes. Under Scene Composition Shift, MaskHarness-WAM completes 11 of 12 trials, achieving 91.7\% SR and 95.8\% PR. In comparison, MaskWAM, Fast-WAM, and $\pi_0$ achieve 41.7\%/70.8\%, 0.0\%/12.5\%, and 8.3\%/25.0\%, respectively. While MaskWAM preserves partial progress in several trials, it is less reliable in maintaining correct target selection and task progression through the full sequence.

Under the more challenging Appearance-and-Geometry Shift, MaskHarness-WAM achieves 83.3\% SR and 83.3\% PR, compared with 58.3\%/79.2\% for MaskWAM, 41.7\%/54.2\% for Fast-WAM, and 0.0\%/16.7\% for $\pi_0$. Overall, the mask-conditioned methods are more robust than the non-mask baselines under visual distribution shifts, indicating the benefit of explicit instance-level spatial grounding. MaskHarness-WAM further improves over MaskWAM by coupling this grounding with closed-loop target switching and task-state management, leading to more reliable long-horizon execution across changes in appearance, scene composition, and geometry.

\subsection{Zero-Shot Generalization}

We further evaluate MaskHarness-WAM in two zero-shot settings without additional training or task-specific fine-tuning: Single-Step Transfer and Unseen-Category Transfer.

In Single-Step Transfer, the robot must manipulate only the language-specified target among four candidate blocks and terminate immediately after completing the operation. This setting evaluates whether a policy trained on four-step sequential tasks can adapt to a single-step instruction with correct target selection and termination. Since the task contains only one subtask, SR and PR are identical. Each method is evaluated over 4 trials.

In Unseen-Category Transfer, the robot must sequentially manipulate four bread instances in the language-specified order. Bread objects are never observed during training, making this setting a direct test of instance-level grounding and long-horizon execution on an unseen object category. Each method is evaluated over 25 trials. Figure~\ref{fig:generalization_scenes}(d) shows the Unseen-Category Transfer setting, and Table~\ref{tab:zeroshot_results} reports the quantitative results.

\begin{table}[htbp]
    \centering
    \caption{Zero-shot generalization under Single-Step Transfer and Unseen-Category Transfer (SR/PR, \%).}
    \label{tab:zeroshot_results}
    \resizebox{\columnwidth}{!}{%
    \begin{tabular}{lcccc}
        \toprule
        \textbf{Zero-Shot Setting} & \textbf{$\boldsymbol{\pi_0}$} & \textbf{Fast-WAM} & \textbf{MaskWAM} & \textbf{MaskHarness-WAM} \\
        \midrule
        \textbf{Single-Step Transfer} 
        & 0.0/0.0 
        & 0.0/0.0 
        & 0.0/0.0 
        & \textbf{100.0/100.0} \\
        
        \textbf{Unseen-Category Transfer} 
        & 0.0/4.2 
        & 4.2/12.5 
        & 16.7/34.4 
        & \textbf{68.0/77.0} \\
        \bottomrule
    \end{tabular}}
\end{table}

Under Single-Step Transfer, MaskHarness-WAM successfully selects the specified target, completes the manipulation, and terminates correctly in all 4 trials, achieving 100.0\% SR and PR. In contrast, none of the three baselines completes the task successfully. This result shows that the harness can adapt the execution structure from multi-step training tasks to a single-step instruction without additional training.

Under Unseen-Category Transfer, MaskHarness-WAM achieves 68.0\% SR and 77.0\% PR, compared with 16.7\%/34.4\% for MaskWAM, 4.2\%/12.5\% for Fast-WAM, and 0.0\%/4.2\% for $\pi_0$. Despite the complete absence of bread objects during training, MaskHarness-WAM maintains reliable instance grounding and ordered task progression across the four-step sequence. The large gap over MaskWAM further indicates that mask-based grounding alone is insufficient for reliable long-horizon transfer; closed-loop task-state management is also critical for converting local manipulation capability into complete sequential execution.

\section{Conclusion}

We introduced \emph{MaskHarness-WAM}, an instance-grounded harness that extends a mask-conditioned world action model to reliable long-horizon robot manipulation without modifying the underlying low-level policy. By combining subtask-specific target masks with task planning, state tracking, visual verification, recovery, and execution scheduling, the harness maintains correct instance grounding and task progression across sequential manipulation stages. Extensive real-robot experiments on cube and test-tube tasks, including out-of-distribution and zero-shot settings, demonstrate that MaskHarness-WAM consistently improves long-horizon execution reliability, highlighting the effectiveness of instance-grounded harnessing for extending finite-horizon manipulation policies to more general robotic tasks.


\bibliographystyle{IEEEtran}
\bibliography{ref}

@inproceedings{carion2026sam,
  title={Sam 3: Segment anything with concepts},
  author={Carion, Nicolas and Gustafson, Laura and Hu, Yuan-Ting and Debnath, Shoubhik and Hu, Ronghang and Suris Coll-Vinent, Didac and Ryali, Chaitanya and Alwala, Kalyan Vasudev and Khedr, Haitham and Huang, Andrew and others},
  booktitle={International conference on learning representations},
  volume={2026},
  pages={138846--138923},
  year={2026}
}

@article{yu2026maskwam,
  title={Maskwam: Unifying mask prompting and prediction for world-action models},
  author={Yu, Hanyang and Lin, Haitao and Zhang, Jingbo and Zhang, Wenyao and Gu, Chenghao and Li, Heng and Tan, Ping},
  journal={arXiv preprint arXiv:2606.13515},
  year={2026}
}

@article{zhang2026harnessvla,
  title={Harness VLA: Steering Frozen VLAs into Reliable Manipulation Primitives via Memory-Guided Agents},
  author={Zhang, Yixian and Zhang, Huanming and Gao, Feng and Li, Xiao and Liu, Zhihao and Zhu, Chunyang and Qiu, Jiaxing and Yan, Yuchen and Liu, Jiyuan and Tang, Wenhao and others},
  journal={arXiv preprint arXiv:2607.08448},
  year={2026}
}

@article{gu2026harnesswam,
  title={HarnessWAM: Bridging Prediction and Deliberation in World Action Models},
  author={Gu, Zhaopeng and Zhu, Bingke and Lin, Tianxi and Zhu, Guibo and Chen, Yingying and Wang, Kai and Yuan, Tingyu and Zhao, Chaoyang and Li, Zhaowen and Su, Peng and others},
  journal={arXiv preprint arXiv:2608.09516},
  year={2026}
}

@article{huang2026roboharness,
  title={RoboHarness: Memory-Driven Orchestration of Heterogeneous Robot Policies for Long-Horizon Planning},
  author={Huang, Jinbang and Hu, Yuanzhao and Li, Zhiyuan and Qi, Ran and Xiao, Yixin and Zhang, Zhanguang and Coates, Mark and Cao, Tongtong and Zhang, Yingxue},
  journal={arXiv preprint arXiv:2607.18060},
  year={2026}
}

@article{wang2026thea,
  title={Towards the Harness of Embodied Agents},
  author={Wang, Qi and Wang, Tianyi and Li, Chengyang and Ban, Shikun and Chen, Yurun and Ge, Yizhong and Qin, Jason and Li, Chengtai and Zhu, Wentao},
  journal={arXiv preprint arXiv:2608.11246},
  year={2026}
}

@article{ding2026zetta,
  title={{Zetta $\zeta$}: An Efficient Closed-Loop Embodied Harness for Self-Evolving Physical Intelligence},
  author={Ding, Xin and Mi, Liang and Huang, Mingzhe and Wang, Zixuan and Zhang, Chao and Hao, Zixu and Chen, Fu and Li, Xiangyu and Zheng, Yikai and Guo, Yaoyu and others},
  journal={arXiv preprint arXiv:2608.16590},
  year={2026}
}

@article{wang2026shaper,
  title={Self-Evolving Embodied Agents via Skill-Harness Evolution},
  author={Wang, Peidong and Ma, Zhiming and Chang, Ying and Luo, Xufang and Yang, Xiaocui and Feng, Shi and Yang, Yuqing and Li, Dongsheng},
  journal={arXiv preprint arXiv:2608.11350},
  year={2026}
}

@article{chen2026showharness,
  title={Show-Harness: Just a VLM Agent Can Play Robots},
  author={Chen, Yanzhe and Bai, Zechen and Cao, Zhijun and Zeng, Wenzheng and Lin, Kevin Qinghong and Lin, Yiqi and Liang, Guoqiang and Ma, Kevin Yuchen and Huang, Qiming and Shou, Mike Zheng},
  journal={arXiv preprint arXiv:2609.10522},
  year={2026}
}

@article{ye2026worldaction,
  title={World action models are zero-shot policies},
  author={Ye, Seonghyeon and Ge, Yunhao and Zheng, Kaiyuan and Gao, Shenyuan and Yu, Sihyun and Kurian, George and Indupuru, Suneel and Tan, You Liang and Zhu, Chuning and Xiang, Jiannan and others},
  journal={arXiv preprint arXiv:2602.15922},
  year={2026}
}

@article{yuan2026fastwam,
  title={Fast-wam: Do world action models need test-time future imagination?},
  author={Yuan, Tianyuan and Dong, Zibin and Liu, Yicheng and Zhao, Hang},
  journal={arXiv preprint arXiv:2603.16666},
  year={2026}
}

@article{motubrain2026,
  title={Motubrain: An advanced world action model for robot control},
  author={Team, MotuBrain and Xiang, Chendong and Bao, Fan and Liu, Haitian and Tan, Hengkai and Bi, Hongzhe and Li, James and Liu, Jiabao and Pang, Jingrui and Jing, Kiro and others},
  journal={arXiv preprint arXiv:2604.27792},
  year={2026}
}

@article{gu2023rttrajectory,
  title={Rt-trajectory: Robotic task generalization via hindsight trajectory sketches},
  author={Gu, Jiayuan and Kirmani, Sean and Wohlhart, Paul and Lu, Yao and Arenas, Montserrat Gonzalez and Rao, Kanishka and Yu, Wenhao and Fu, Chuyuan and Gopalakrishnan, Keerthana and Xu, Zhuo and others},
  journal={arXiv preprint arXiv:2311.01977},
  year={2023}
}

@article{fang2024moka,
  title={Moka: Open-world robotic manipulation through mark-based visual prompting},
  author={Liu, Fangchen and Fang, Kuan and Abbeel, Pieter and Levine, Sergey},
  journal={arXiv preprint arXiv:2403.03174},
  year={2024}
}

@article{li2025controlvla,
  title={Controlvla: Few-shot object-centric adaptation for pre-trained vision-language-action models},
  author={Li, Puhao and Wu, Yingying and Xi, Ziheng and Li, Wanlin and Huang, Yuzhe and Zhang, Zhiyuan and Chen, Yinghan and Wang, Jianan and Zhu, Song-Chun and Liu, Tengyu and others},
  journal={arXiv preprint arXiv:2506.16211},
  year={2025}
}

@inproceedings{huang2025roboground,
  title={Roboground: Robotic manipulation with grounded vision-language priors},
  author={Huang, Haifeng and Chen, Xinyi and Chen, Yilun and Li, Hao and Han, Xiaoshen and Wang, Zehan and Wang, Tai and Pang, Jiangmiao and Zhao, Zhou},
  booktitle={2025 IEEE/CVF Conference on Computer Vision and Pattern Recognition (CVPR)},
  pages={22540--22550},
  year={2025},
  organization={IEEE}
}

@inproceedings{liang2023codeaspolicies,
  title={Code as policies: Language model programs for embodied control},
  author={Liang, Jacky and Huang, Wenlong and Xia, Fei and Xu, Peng and Hausman, Karol and Ichter, Brian and Florence, Pete and Zeng, Andy},
  booktitle={2023 IEEE International conference on robotics and automation (ICRA)},
  pages={9493--9500},
  year={2023},
  organization={IEEE}
}

@article{singh2023progprompt,
  title={Progprompt: Generating situated robot task plans using large language models},
  author={Singh, Ishika and Blukis, Valts and Mousavian, Arsalan and Goyal, Ankit and Xu, Danfei and Tremblay, Jonathan and Fox, Dieter and Thomason, Jesse and Garg, Animesh},
  journal={arXiv preprint arXiv:2209.11302},
  year={2022}
}

@article{du2023unipi,
  title={Learning universal policies via text-guided video generation},
  author={Du, Yilun and Yang, Sherry and Dai, Bo and Dai, Hanjun and Nachum, Ofir and Tenenbaum, Josh and Schuurmans, Dale and Abbeel, Pieter},
  journal={Advances in neural information processing systems},
  volume={36},
  pages={9156--9172},
  year={2023}
}

@inproceedings{wu2023gr1,
  title={Unleashing large-scale video generative pre-training for visual robot manipulation},
  author={Wu, Hongtao and Jing, Ya and Cheang, Chilam and Chen, Guangzeng and Xu, Jiafeng and Li, Xinghang and Liu, Minghuan and Li, Hang and Kong, Tao},
  booktitle={International Conference on Learning Representations},
  volume={2024},
  pages={10641--10662},
  year={2024}
}

@article{cheang2024gr2,
  title={Gr-2: A generative video-language-action model with web-scale knowledge for robot manipulation},
  author={Cheang, Chi-Lam and Chen, Guangzeng and Jing, Ya and Kong, Tao and Li, Hang and Li, Yifeng and Liu, Yuxiao and Wu, Hongtao and Xu, Jiafeng and Yang, Yichu and others},
  journal={arXiv preprint arXiv:2410.06158},
  year={2024}
}

@article{bharadhwaj2024gen2act,
  title={Gen2act: Human video generation in novel scenarios enables generalizable robot manipulation},
  author={Bharadhwaj, Homanga and Dwibedi, Debidatta and Gupta, Abhinav and Tulsiani, Shubham and Doersch, Carl and Xiao, Ted and Shah, Dhruv and Xia, Fei and Sadigh, Dorsa and Kirmani, Sean},
  journal={arXiv preprint arXiv:2409.16283},
  year={2024}
}

@article{zhou2024robodreamer,
  title={Robodreamer: Learning compositional world models for robot imagination},
  author={Zhou, Siyuan and Du, Yilun and Chen, Jiaben and Li, Yandong and Yeung, Dit-Yan and Gan, Chuang},
  journal={arXiv preprint arXiv:2404.12377},
  year={2024}
}

@article{zhu2025uwm,
  title={Unified world models: Coupling video and action diffusion for pretraining on large robotic datasets},
  author={Zhu, Chuning and Yu, Raymond and Feng, Siyuan and Burchfiel, Benjamin and Shah, Paarth and Gupta, Abhishek},
  journal={arXiv preprint arXiv:2504.02792},
  year={2025}
}

@article{zhang2025dreamvla,
  title={Dreamvla: a vision-language-action model dreamed with comprehensive world knowledge},
  author={Zhang, Wenyao and Liu, Hongsi and Qi, Zekun and Wang, Yunnan and Yu, Xinqiang and Zhang, Jiazhao and Dong, Runpei and He, Jiawei and Wang, He and Zhang, Zhizheng and others},
  journal={Advances in Neural Information Processing Systems},
  volume={38},
  pages={24195--24228},
  year={2026}
}

@article{ma2026dit4dit,
  title={Dit4dit: Jointly modeling video dynamics and actions for generalizable robot control},
  author={Ma, Teli and Zheng, Jia and Wang, Zifan and Jiang, Chunli and Cui, Andy and Liang, Junwei and Yang, Shuo},
  journal={arXiv preprint arXiv:2603.10448},
  year={2026}
}

@article{stone2023moo,
  title={Open-world object manipulation using pre-trained vision-language models},
  author={Stone, Austin and Xiao, Ted and Lu, Yao and Gopalakrishnan, Keerthana and Lee, Kuang-Huei and Vuong, Quan and Wohlhart, Paul and Kirmani, Sean and Zitkovich, Brianna and Xia, Fei and others},
  journal={arXiv preprint arXiv:2303.00905},
  year={2023}
}

@article{sundaresan2023kite,
  title={Kite: Keypoint-conditioned policies for semantic manipulation},
  author={Sundaresan, Priya and Belkhale, Suneel and Sadigh, Dorsa and Bohg, Jeannette},
  journal={arXiv preprint arXiv:2306.16605},
  year={2023}
}

@article{haldar2025pointpolicy,
  title={Point policy: Unifying observations and actions with key points for robot manipulation},
  author={Haldar, Siddhant and Pinto, Lerrel},
  journal={arXiv preprint arXiv:2502.20391},
  year={2025}
}

@article{huang2024rekep,
  title={Rekep: Spatio-temporal reasoning of relational keypoint constraints for robotic manipulation},
  author={Huang, Wenlong and Wang, Chen and Li, Yunzhu and Zhang, Ruohan and Fei-Fei, Li},
  journal={arXiv preprint arXiv:2409.01652},
  year={2024}
}

@inproceedings{bharadhwaj2024track2act,
  title={Track2act: Predicting point tracks from internet videos enables generalizable robot manipulation},
  author={Bharadhwaj, Homanga and Mottaghi, Roozbeh and Gupta, Abhinav and Tulsiani, Shubham},
  booktitle={European Conference on Computer Vision},
  pages={306--324},
  year={2024},
  organization={Springer}
}

@article{wen2023atm,
  title={Any-point trajectory modeling for policy learning},
  author={Wen, Chuan and Lin, Xingyu and So, John and Chen, Kai and Dou, Qi and Gao, Yang and Abbeel, Pieter},
  journal={arXiv preprint arXiv:2401.00025},
  year={2023}
}

@inproceedings{bahl2023vrb,
  title={Affordances from human videos as a versatile representation for robotics},
  author={Bahl, Shikhar and Mendonca, Russell and Chen, Lili and Jain, Unnat and Pathak, Deepak},
  booktitle={2023 IEEE/CVF Conference on Computer Vision and Pattern Recognition (CVPR)},
  pages={01--13},
  year={2023},
  organization={IEEE}
}

@inproceedings{nasiriany2025rtaffordance,
  title={Rt-affordance: Affordances are versatile intermediate representations for robot manipulation},
  author={Nasiriany, Soroush and Kirmani, Sean and Ding, Tianli and Smith, Laura and Zhu, Yuke and Driess, Danny and Sadigh, Dorsa and Xiao, Ted},
  booktitle={2025 IEEE International Conference on Robotics and Automation (ICRA)},
  pages={8249--8257},
  year={2025},
  organization={IEEE}
}

@article{black2025pi0,
  title={{$\pi_0$}: A Vision-Language-Action Flow Model for General Robot Control},
  author={Black, Kevin and Brown, Noah and Driess, Danny and Esmail, Adnan and Equi, Michael and Finn, Chelsea and Fusai, Niccolo and Groom, Lachy and Hausman, Karol and Ichter, Brian and others},
  journal={arXiv preprint arXiv:2410.24164},
  year={2024}
}

@article{huang2023voxposer,
  title={Voxposer: Composable 3d value maps for robotic manipulation with language models},
  author={Huang, Wenlong and Wang, Chen and Zhang, Ruohan and Li, Yunzhu and Wu, Jiajun and Fei-Fei, Li},
  journal={arXiv preprint arXiv:2307.05973},
  year={2023}
}

@article{ahn2022saycan,
  title={Do as i can, not as i say: Grounding language in robotic affordances},
  author={Ahn, Michael and Brohan, Anthony and Brown, Noah and Chebotar, Yevgen and Cortes, Omar and David, Byron and Finn, Chelsea and Fu, Chuyuan and Gopalakrishnan, Keerthana and Hausman, Karol and others},
  journal={arXiv preprint arXiv:2204.01691},
  year={2022}
}

@article{huang2022innermonologue,
  title={Inner monologue: Embodied reasoning through planning with language models},
  author={Huang, Wenlong and Xia, Fei and Xiao, Ted and Chan, Harris and Liang, Jacky and Florence, Pete and Zeng, Andy and Tompson, Jonathan and Mordatch, Igor and Chebotar, Yevgen and others},
  journal={arXiv preprint arXiv:2207.05608},
  year={2022}
}

@article{rana2023sayplan,
  title={Sayplan: Grounding large language models using 3d scene graphs for scalable robot task planning},
  author={Rana, Krishan and Haviland, Jesse and Garg, Sourav and Abou-Chakra, Jad and Reid, Ian and Suenderhauf, Niko},
  journal={arXiv preprint arXiv:2307.06135},
  year={2023}
}

@article{yang2025qwen3,
  title={Qwen3 technical report},
  author={Yang, An and Li, Anfeng and Yang, Baosong and Zhang, Beichen and Hui, Binyuan and Zheng, Bo and Yu, Bowen and Gao, Chang and Huang, Chengen and Lv, Chenxu and others},
  journal={arXiv preprint arXiv:2505.09388},
  year={2025}
}

@article{su2026demavla,
  title={DeMaVLA: A Vision-Language-Action Foundation Model for Generalizable Deformable Manipulation},
  author={Su, Taiyi and Zhu, Jian and Wang, Tianjian and He, Youzhang and Huang, Zitai and Zhang, Jianjun and Ma, Chong and Wang, Hanyang and Zhang, Tianjiao and Yin, Munan and others},
  journal={arXiv preprint arXiv:2605.31286},
  year={2026}
}

@article{zhu2026dswam,
  title={DSWAM: A Dual-System World Action Foundation Model for Fine-Grained Robot Manipulation},
  author={Zhu, Jian and Zhang, Jianjun and Su, Taiyi and Liu, Tianbin and Wang, Zhangyuan and Xie, Kai and Huang, Zitai and Ma, Chong and He, Youzhang and Wang, Tianjian and others},
  journal={arXiv preprint arXiv:2607.04927},
  year={2026}
}

@article{zhang2026learning,
  title={Learning 4D Geometric Priors for Inference-Efficient World Action Models},
  author={Zhang, Jianjun and Zhu, Jian and Su, Taiyi and Ma, Chong and Huang, Zitai and Xu, Yi and Wang, Hanli},
  journal={arXiv preprint arXiv:2607.05468},
  year={2026}
}

\end{document}